# Using Sentiment and Technical Analysis to Predict Bitcoin with Machine Learning

Arthur Emanuel de Oliveira Carosia[1]

Federal Institute of São Paulo, São João da Boa Vista, Brazil

**Abstract.** *Cryptocurrencies have gained significant attention in recent years due to their decentralized nature and potential for financial innovation. Thus, the ability to accurately predict its price has become a subject of great interest for investors, traders, and researchers. Some works in the literature show how Bitcoin's market sentiment correlates with its price fluctuations in the market. However, papers that consider the sentiment of the market associated with financial Technical Analysis indicators in order to predict Bitcoin's price are still scarce. In this paper, we present a novel approach for predicting Bitcoin price movements by combining the Fear & Greedy Index, a measure of market sentiment, Technical Analysis indicators, and the potential of Machine Learning algorithms. This work represents a preliminary study on the importance of sentiment metrics in cryptocurrency forecasting. Our initial experiments demonstrate promising results considering investment returns, surpassing the Buy & Hold baseline, and offering valuable insights about the combination of indicators of sentiment and market in a cryptocurrency prediction model.*

## 1. Introduction

Bitcoin was introduced in 2008 by an anonymous entity known as Satoshi Nakamoto (Nakamoto, 2008). It utilizes cryptographic techniques to secure transactions, verify the transfer of digital assets, and maintain a transparent and immutable ledger. One of the main Bitcoin advantages is that it operates on a decentralized network called blockchain, reducing the dependence on central authorities and promoting a more transparent and resilient financial system. Bitcoin utilizes advanced cryptographic techniques, offering robust security and protecting users' financial privacy, preventing unauthorized access and censorship. Besides, Bitcoin's limited supply and scarcity properties make it an attractive store of value, protecting against inflation and providing an alternative to traditional asset classes (Tschorsch & Scheuermann, 2016).

---

[1]    Contact: arthuremanuel.carosia@ifsp.edu.br

However, over the years, Bitcoin has experienced substantial price fluctuations (Ghimire & Selvaraj, 2019), driven by factors such as market demand, regulatory developments, macroeconomic indicators, investor sentiment, and technological advancements. Besides, the volatility and rapid price changes in the Bitcoin market pose challenges and opportunities for investors and traders. Accurate price prediction models can provide valuable insights into market behavior, aiding investors in making informed decisions regarding buying, selling, or holding Bitcoin. Additionally, financial institutions, regulators, and policymakers can benefit from a better understanding of the factors driving Bitcoin price movements.

Traditional investors often consider two different approaches in order to analyze the stock market: Technical and Fundamental Analysis (Achelis, 2000). Technical Analysis primarily focuses on studying historical market data to identify patterns, trends, and potential price movements. It relies on the belief that past market behavior can predict future movements, without much emphasis on the underlying value of the asset or the company's financials. On the other hand, Fundamental Analysis focuses on evaluating the intrinsic value of an asset by examining the underlying factors of a company, like its financial statements, earnings, growth prospects, and management team, among others (Kumbure et al., 2022a). It aims to determine whether an asset is overvalued or undervalued in the market, regardless of short-term price fluctuations. Besides, Fundamental Analysis, in the context of behavioral finance, can also consider the sentiment, emotions, and perceptions of investors and market participants towards a particular asset, industry, or market in order to decide to invest in a given stock or market (Chu et al., 2019). Market sentiment is crucial as it provides valuable insights into the collective emotions, attitudes, and expectations of market participants, influencing investment decisions and market trends. Besides, it is worth highlighting that both Technical and Fundamental analysis have been used in the literature in order to study Bitcoin.

More recently, Machine Learning, a branch of Artificial Intelligence, has shown remarkable success in various domains, including finance, surpassing traditional market predicting and analysis techniques (Nosratabadi et al., 2020). By leveraging historical price data, market indicators, and other relevant features, Machine Learning algorithms can learn non-linear patterns and relationships to make predictions. In the context of Bitcoin, machine learning techniques offer the potential to uncover hidden patterns and factors influencing price movements, aiding in the development of accurate prediction models, as recent works show (Nazareth & Ramana Reddy, 2023).

Besides, the literature on Bitcoin prediction using Machine Learning techniques often relies on historical data: open, high, low, and close prices alongside volume (Nazareth & Ramana Reddy, 2023). This work, however, considers, besides historical data, Technical and Fundamental Analysis data. It is still scarce in the literature works that consider elements from both analyses in cryptocurrency prediction.

Thus, this work aims to predict the Bitcoin future price using Machine Learning algorithms, technical indicators, from Technical Analysis, and the Fear & Greed Index, which measures Bitcoin's market sentiment, from Fundamental Analysis. The following Machine Learning algorithms were considered: Linear Regression, Support Vector Machines, XGBoost, Gradient Boosting, Random Forests, and Multilayer Perceptron. Our results showed that the combination of technical indicators and sentiment measure overcomes the profit when considering the Buy & Hold investment strategy. The insights gained from this study can potentially enhance investment strategies, risk management approaches, and policy-making decisions in the context of Bitcoin and other cryptocurrencies.

The remainder of this paper is organized as follows. Section 2 presents the theoretical foundation and related works. Section 3 presents the Methodology, and Section 4 presents the Results. Finally, Section 5 concludes the work and presents the limitations and future works.

## 2. Theoretical Foundation

Financial market forecasting is a task widely studied in academic literature (Kumbure et al., 2022b), while cryptocurrency prediction is a recent field of study (Nazareth & Ramana Reddy, 2023). In this sense, many works and methods have been developed in order to predict the future prices of a given asset, and, in the last decade, the main highlight has been the use of Machine Learning and Data Science techniques (Picasso et al., 2019).

This Section presents the theoretical foundation necessary for the development of this work and is organized as follows. Section 2.1 presents the Machine Learning algorithms selected for this work and their description. Section 2.2 presents the related works, that is, studies that deal with the topic of cryptocurrency prediction using Machine Learning algorithms, as well as presenting the differences between this work and its counterparts.

### 2.1 Machine Learning Techniques

There are several Machine Learning algorithms that are used in the financial market and cryptocurrency prediction literature (Duarte et al., 2020). Among the most used algorithms, it is worth highlighting: Linear Regression, Support Vector Machines, Random Forest, XGBoost, Gradient Boosting, and Multilayer Perceptron (Mujlid, 2023; Nazareth & Ramana Reddy, 2023). A brief description of each of them is presented in the following.

**Linear Regression**: technique based on the creation of a linear model with the objective of minimizing the sum of the residual of squares between the elements that are observed in the data set and the elements that are predicted by the linear approximation (Su et al., 2012).

**Support Vector Machines**: technique whose operation consists of maximizing the margin of class separation through support vectors. This technique also presents the possibility of using kernel functions for non-linear data, making it easier to find the limit of separation between classes. Furthermore, it is worth highlighting that the Support Vector Machines technique is used for classification tasks. In this work, we use its adaptation for regression tasks, called Support Vector Regression (Noble, 2006).

**Random Forest**: a decision tree-based algorithm that uses a technique called bagging, in which each tree in a "forest" of trees is trained independently (Breiman, 2001).

**Gradient Boosting**: a technique that sequentially adds predictors to an ensemble, each one correcting its predecessor. This method tries to fit the new predictor to the residual errors made by the previous predictor (Duarte et al., 2020).

**XGBoost**: tree-based technique, which consists of the latest version of the Gradient Boosting algorithm, featuring improvements to deal with sparse data and data regularization capabilities (T. Chen & Guestrin, n.d.).

**Multilayer Perceptron**: technique consisting of an Artificial Neural Network in which each neuron, called Perceptron, is followed by an activation function and organized in layers. Each neuron receives impulses from the entire previous layer and propagates a new impulse based on its activation function during training (Gardner & Dorling, 1998).

## 2.2 Related Works

Many works use Machine Learning methods to predict the future value of assets. Comprehensive reviews of several works in the literature can be found in (Nazareth & Ramana Reddy, 2023), (Kumbure et al., 2022b), (Khedr et al., 2021), and (Henrique et al., 2019), which highlight that the most common works in the area normally use traditional Machine Learning models, such as Support Vector Machines (SVMs) and Linear Regression, or different types of Artificial Neural Networks (ANNs). Below, the most relevant works to this study are detailed.

Lahmiri and Bekiros (2019) used Artificial Neural Network methods to predict the future price of cryptocurrencies, including Bitcoin, and their findings show that the Long Short-Term Memory (LSTM) neural network model has a better performance in crypto prediction. Altan et al. (2019) also show that integrating LSTM and empirical wavelet transform (EWT) improves the performance of cryptocurrency price prediction. Jiang and Liang (2018) used a Convolutional Neural Network model to predict Bitcoin's future price. The authors trained their proposed model using historical data and also presented an investment simulation in which the proposal didn't present a better performance than the baselines.

Z. Chen et al. (2020) compared linear models and several traditional Machine Learning algorithms, showing that linear models achieved superior performance in cryptocurrency prediction when considering low-frequency data. Also, they presented that the SVM technique outperformed other Machine Learning models when predicting the Bitcoin price. Saad et al. (2020) studied Bitcoin and Ethereum prediction with the use of Machine Learning methods, such as Linear Regression, Random Forest, and Gradient Boosting. The results showed an accuracy of up to 99% for Bitcoin and Ethereum price prediction. Poongodi et al. (2020) examined linear models and the SVM model for predicting the price of the cryptocurrency Ethereum, showing that the SVM model outperforms the Linear Regression model considering the accuracy measure.

Considering the related works in the literature, this paper presents the following contributions: (1) the use of several traditional Machine Learning algorithms for comparison with an Artificial Neural Network-based algorithm; and (2) the use of market indicators alongside sentiment indicators as input features for the Machine Learning models considered.

## 3. Methodology

This work aims to predict Bitcoin's future price considering the following features: Bitcoin's market value, Technical Indicators, and a measure of the market's sentiment. To this, we propose the following steps, illustrated in Figure 1:

1. Preprocessing: in this step, the data are downloaded and combined in one feature vector, composed of three days of Bitcoin's price, Technical Indicators, and market sentiment. These data are normalized before being used as input by the Machine Learning algorithms.
2. Prediction and Evaluation: in this step, the feature vector is used as input in the considered Machine Learning algorithms in order to predict the next day of Bitcoin's price. Then, the models are evaluated considering their internal parameters and prediction results.
3. Investment Simulation: in this final step, the models selected in the previous step are used to perform an investment simulation to evaluate the proposal in a real environment. Besides, in this step, we also compared the predictions with the investment strategy Buy & Hold.

The proposed steps are detailed in the following sections.

## 3.1 Dataset

The dataset used in the experiments consists of: (1) daily quotations of Bitcoin's traded value in US Dollars, which are freely available on Yahoo Finance (https://finance.yahoo.com/); and (2) the Crypto Fear & Greedy index, which is the measure of daily emotions and sentiments from different sources in the internet, which is also freely available (https://alternative.me/crypto/fear-and-greed-index/). The dataset comprises the period from 01-Fev-2018 to 01-Jan-2023.

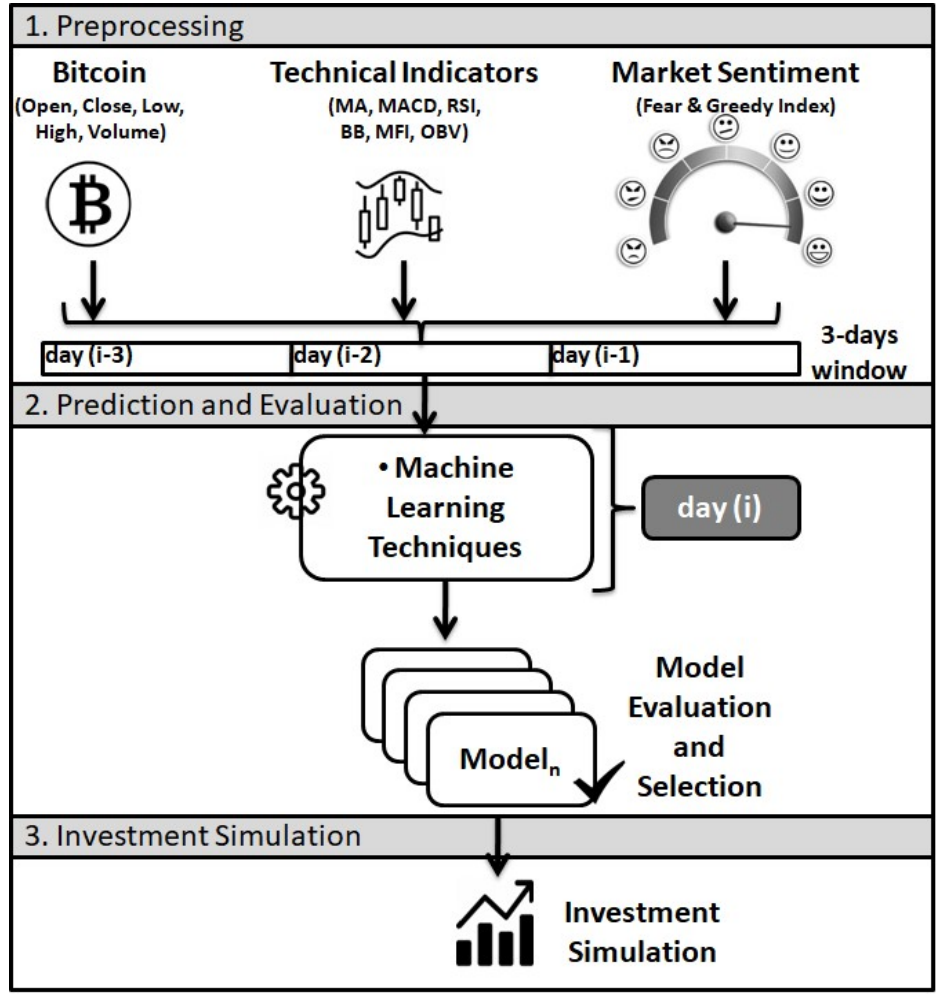

**Figure 1.** Proposed methodology. First, the data are collected and combined in a feature vector. In the following, the feature vector is used to predict Bitcoin's price. Finally, the predictions are evaluated and the created models are used to perform an investment simulation.

Bitcoin's quotation data are composed of the following features: opening, maximum, minimum, closing, and volume. Figure 2 illustrates the behavior of Bitcoin's value in the considered period, while the distribution of the data used in the experiments is presented in Table 1.

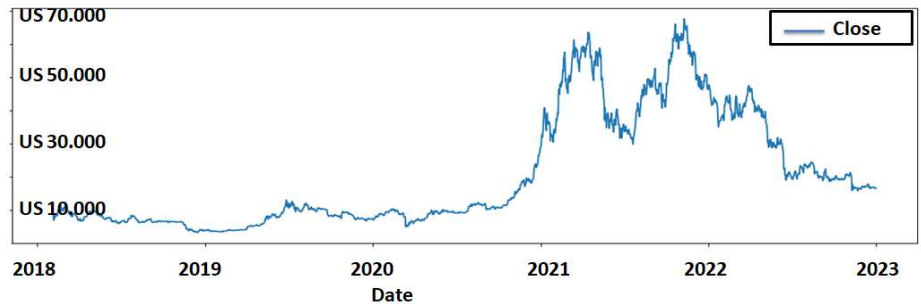

**Figure 2.** Bitcoin's valuation considering the period of 01-Fev-2018 to 01-Jan-2023

**Table 1.** Data distribution

|  | **Open** | **High** | **Low** | **Close** | **Volume** |
|---|---|---|---|---|---|
| **Number** | 1953 | 1953 | 1953 | 1953 | 1953 |
| **Mean** | 20877,1 | 21374,10 | 20328,95 | 20883,67 | 2,639E+10 |
| **STD** | 16471,6 | 16898,94 | 15974,53 | 16465,56 | 1,931E+10 |
| **Min** | 3236,2? | 3275,38 | 3191,30 | 3236,76 | 2,924E+09 |
| **25%** | 7989,3? | 8196,65 | 7786,05 | 7987,37 | 1,367E+10 |
| **50%** | 11913,0 | 12150,99 | 11681,48 | 11941,97 | 2,41E+10 |
| **75%** | 31533,8 | 32564,03 | 30044,50 | 31533,07 | 3,517E+10 |
| **Max** | 67549,7 | 68789,63 | 66382,06 | 67566,83 | 3,51E+11 |

Finally, the Crypto Fear & Greedy Index (available at https://www.binance.com/pt-BR/square/fear-and-greed-index or at https://alternative.me/crypto/fear-and-greed-index/) consists of a number that summarizes the market sentiment, where a value of 0 means "Extreme Fear" while a value of 100 represents "Extreme Greed". The index is calculated considering five data sources: (1) Bitcoin's volatility in the market; (2) social media analysis in Twitter; (3) surveys asking people how they are seeing the market; (4) dominance of the coin in the whole crypto market; and (5) trends in search engines like Google. The daily value of this index in the 2018-2023 period is presented in Figure 3.

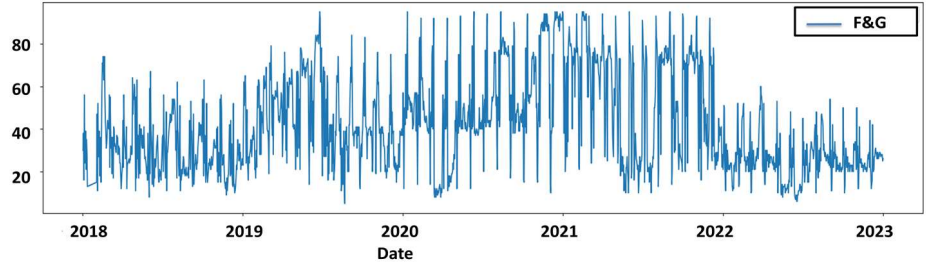

**Figure 3.** Crypto Fear & Greedy Index (F&G) values in the period Fev-2018 to Jan-2013

## 3.1 Preprocessing

Once the data is obtained, it is used to calculate Technical Indicators, which are values that present market trends to the investor. The following indicators were used in the experiments (Li et al., 2020): Moving Average (MA), with the parameters 10, 20, and 30 days; Moving Average Convergence Divergence (MACD), with the parameters 12, 26, and 9 days; Relative Strength Index, with the parameters 6, 12, and 24 days; Money Flow Index, with the 14-days parameter; On Balance Volume (OBV), and Bollinger Bands (BB), both with no additional parameters.

After the Technical Indicators generation, a feature vector is created containing a 3-day period (that is, day *i-3*, day *i-2*, day *i-1*) in order to predict the closing price of the day *i*. The feature vector is normalized using a min-max approach.

## 3.2 Prediction and Evaluation

In order to predict the next day's closing price of Bitcoin, in this work, we consider the following Machine Learning algorithms, which are commonly used in the prediction literature (Nazareth & Ramana Reddy, 2023): Linear Regression, Support Vector Machines, XGBoost, Gradient Boosting, Random Forest, and Multilayer Perceptron.

As presented before, the feature vector used as input to these algorithms is composed of a 3-day combination of stock prices, Technical Indicators, and the Crypto Fear & Greedy Index. Since most of the Machine Learning algorithms have different parameters, we also performed a grid search considering a set of parameters for each one of them, as presented in Table 2.

Besides, the algorithms were trained with an increasing window cross-validation approach, with 3 steps (Li et al., 2020), in order to maintain the temporal dependence within the stock prices' time series. In this approach, a window is used with a growing number of samples in the training set on each step, while this method reserves a subset of the next data samples as a validation set.

Finally, to evaluate and select the model with the most suitable parameters for the task of predicting the future Bitcoin's price, the Mean Squared Error was used (MSE), since it is widely used in works that perform regression.

**Table 2.** Parameters for each Machine Learning algorithm

| Algorithm | Parameters |
|---|---|
| **Multilayer Perceptron** | max_iter: {250,500,1000}, learning_rate_init: {0.01,0.001}, hidden_layer_sizes: {(10,10,10),(25,25,25),(50,50,50), (10,10,10,10,10),(25,25,25,25,25),(50,50,50,50,50) } early_stopping = true |
| **XG Boost** | booster: {gbtree, gblinear, dart}, max_delta_step: {0,1,5}, lambda:{1,3,5,10,50,100} |
| **Gradient Boosting** | criterion: {friedman_mse}, n_estimators: {150}, learning_rate: {0.001, 0.01, 0.1}, max_depth: {3, 5, 10}, max_leaf_nodes : {5, 10, 35, None}, min_samples_leaf : {1, 3, 5} |
| **Random Forest** | criterion: {squared_error,poisson}, n_estimators: {150}, max_leaf_nodes: {5,10,35,None},min_samples_leaf: {1,3,5} |
| **Support Vector Regression** | C : {0.001, 0.01, 0.1, 1}Kernel: {Linear, RBF} |
| **Linear Regression** | - |

## 3.3 Investment Simulation

To evaluate the most suitable trained models, an investment simulation was performed considering the initial value of USD 200,000.00. The investment strategy used is presented in the following: (1) a prediction p of the closing value of the stock on day i+1 is performed; (2) if p is greater than or equal to 0, the entire available budget is used to buy the cryptocurrency or, if the investor already owns it, he keeps it in his possession; and (3) if p less than 0, all the cryptocurrencies that the investor owns are sold or, if he/she does not own the cryptocurrency, the investor does not take any action.

As a baseline for this experiment, we consider the Buy & Hold investment strategy, which consists of the investor buying the cryptocurrency at the beginning of the investment period and keeping it in his possession throughout the considered period. This strategy is very common among Bitcoin investors.

## 4. Results

The experiments of this work are separated into two steps: (1) the selection of the most suitable model's parameters according to the calculation of the training error metrics, presented in Section 4.1; and (2) the investment simulation, in which the designed model is applied in the financial context, presented in Section 4.2. Finally, Section 4.3 presents a discussion of the obtained results.

### 4.1 Training Metrics

In order to train each one of the proposed Machine Learning algorithms, in this work we consider the period between 01-Fev-2018 to 31-May-2022 as a training set, while the period comprised between 01-Jun-2022 to 31-Dec-2023 is used as a test set. The training methodology used is increasing window cross-validation, as presented in Section 3.3.

Table 3 provides information on the most suitable parameters for all the considered Machine Learning models, that is, the parameters that led the models to the minimum error on the training step. The parameters were determined using a grid-search approach, presented in Section 3.3.

**Table 3.** Most suitable parameters for each considered Machine Learning algorithm

| Model | Parameters (Suitable) |
|---|---|
| Multilayer Perceptron | {'hidden_layer_sizes': (10, 10), 'learning_rate_init': 0.001, 'max_iter': 1000} |
| XGBoost | {'booster': 'dart', 'lambda': 100, 'max_delta_step': 0} |
| Gradient Boosting | {'criterion': 'friedman_mse', 'learning_rate': 0.1, 'max_depth': 3, 'max_leaf_nodes': 5, 'min_samples_leaf': 3, 'n_estimators': 150} |
| RandForest | {'criterion': 'poisson', 'max_leaf_nodes': 35, 'min_samples_leaf': 3, 'n_estimators': 150} |
| Support Vector Regression – Kernel Linear | {'C': 0.1} |
| Support Vector Regression – Kernel RBF | {'C': 1} |

Finally, Table 4 presents the Mean Squared Error (MSE) values for the Machine Learning models on the validation set. The MSE is a metric used to evaluate the performance of regression models, where lower values indicate better predictive performance.

According to these results, the models with the parameters shown in Table 3 were selected for the remaining experiments of this work, that is, the investment simulation presented in the next section.

**Table 4.** Mean Squared Error (MSE) for each considered Machine Learning algorithm.

| Model | MSE |
|---|---|
| Linear Regression | 0,00013 |
| Multilayer Perceptron | 0,00033 |
| Support Vector Regression (RBF) | 0,00043 |
| RandForest | 0,00192 |
| Support Vector Regression (Linear) | 0,00198 |
| Gradient Boosting | 0,00201 |
| XG Boost | 0,00439 |

## 4.2 Investment Simulation

The developed models in this work were used in an investment simulation considering the 6-month period between 01-Jun-2022 and 31-Dec-2022. As a baseline for this experiment, we considered the Buy & Hold investment strategy, which consists of the investor buying the considered asset at the beginning of the investment period and keeping it under possession during the entire period.

Figure 4 shows the investment simulation graphs for each one of the proposed algorithms, while Table 5 provides the investment metrics for each one of these models as well as the Buy & Hold investment strategy. The models are evaluated based on hits, that is, correct predictions on ups and downs movements of the cryptocurrency in the market, and the final investment value considering an initial investment of USD 200,000,00.

In the experiments, the model Support Vector Regression with RBF kernel achieved 103 hits of successful predictions. It generated USD 192,098.47, which represents the final value of the investment based on the model's predictions. It's worth highlighting that some models, such as Random Forest, presented a higher number of hits, however, its final value of investment is lower than the Support Vector Regression with RBF kernel.

Comparatively, the Buy & Hold strategy achieved a value of USD 120,515.66, indicating a significant decline compared to the Machine Learning models in both approaches.

**Table 5.** Investment simulation results considering both Machine Learning algorithms and Buy & Hold strategy.

| Model | Hits | Final Value |
|---|---|---|
| Support Vector Regression- RBF | 103 | USD 192.098,47 |
| Random Forest | 108 | USD 189.131,26 |
| Gradient Boosting | 109 | USD 187.333,91 |
| Multilayer Perceptron | 102 | USD 187.099,19 |
| Support Vector Regression - Linear | 95 | USD 186.752,61 |
| Linear Regression | 98 | USD 186.452,75 |
| XG Boost | 106 | USD 183.197,72 |
| **Average** | 103 | USD 187.437,99 |
| **Buy & Hold** | - | USD 120.515,66 |

### 4.3 Discussion

According to our results, it is worth highlighting the importance of analyzing the functioning of the models in practice, that is, to perform investment simulation. For example, the Linear Regression model presented the smallest error measure for the validation data set. However, this model did not present satisfactory results in the investment simulation. On the other hand, the Multilayer Perceptron (MLP) and the Support Vector Regression (SVR) models presented, respectively, the second and third smallest values among the error measures and, in the simulation of investment, showed significant results when compared to the Linear Regression model. This result may have been obtained because the Linear Regression model presents linear features, which does not occur with models such as MLP, or SVR. In fact, the behavior of the financial market is also non-linear. Therefore, these results show that not all models that present a lower error value during training also present better results when applied to an investment simulation.

Our results also show that Machine Learning models are capable of overcoming the frequently used investment strategy Buy & Hold when considering Bitcoin as an investment. Thus, algorithms such as SVR and Random Forest are highly recommended to investors and should be considered in investment strategies.

Cryptocurrency markets are highly volatile and influenced by a wide range of factors, including regulatory changes, market news, and investor sentiment. As stated by the Adaptive Market Hypothesis (Lo, 2004), investors are often irrational and present unpredictable behavior, thus it's important to consider sentiment measures as a prediction feature.

Furthermore, it is important to highlight that this study has certain limitations that should be mitigated in future work. Thus, it is necessary to emphasize that past financial returns do not reflect future financial returns and, furthermore, different analysis periods may bring different results in terms of returns. Finally, we also highlight that this is a preliminary study considering sentiment measures in cryptocurrency forecasting using Machine Learning. The promising results obtained in this work encourage us to investigate this problem more deeply in the future.

## 5. Conclusion

This paper aims to propose a combination of a crypto market sentiment index and technical indicators in order to feed Machine Learning algorithms in order to predict the one-day ahead of Bitcoin's closing price. Many works in the stock market field already noticed that the investor's sentiment are related to the stock market movement (Chu et al., 2019). However, in the cryptocurrency literature, there are few studies in this sense (Nazareth & Ramana Reddy, 2023).

The obtained results showed that the algorithm Support Vector Regression presented the best results among the analyzed models considering the final value of the investment, while the algorithm Gradient Boosting presented better accuracy in the prediction of the market movement.

In future works, we intend to expand the experiments, considering: (1) Large Language Models; (2) other cryptocurrencies; (3) the use of Deep Learning models as predictors; and (4) different periods in our analysis.

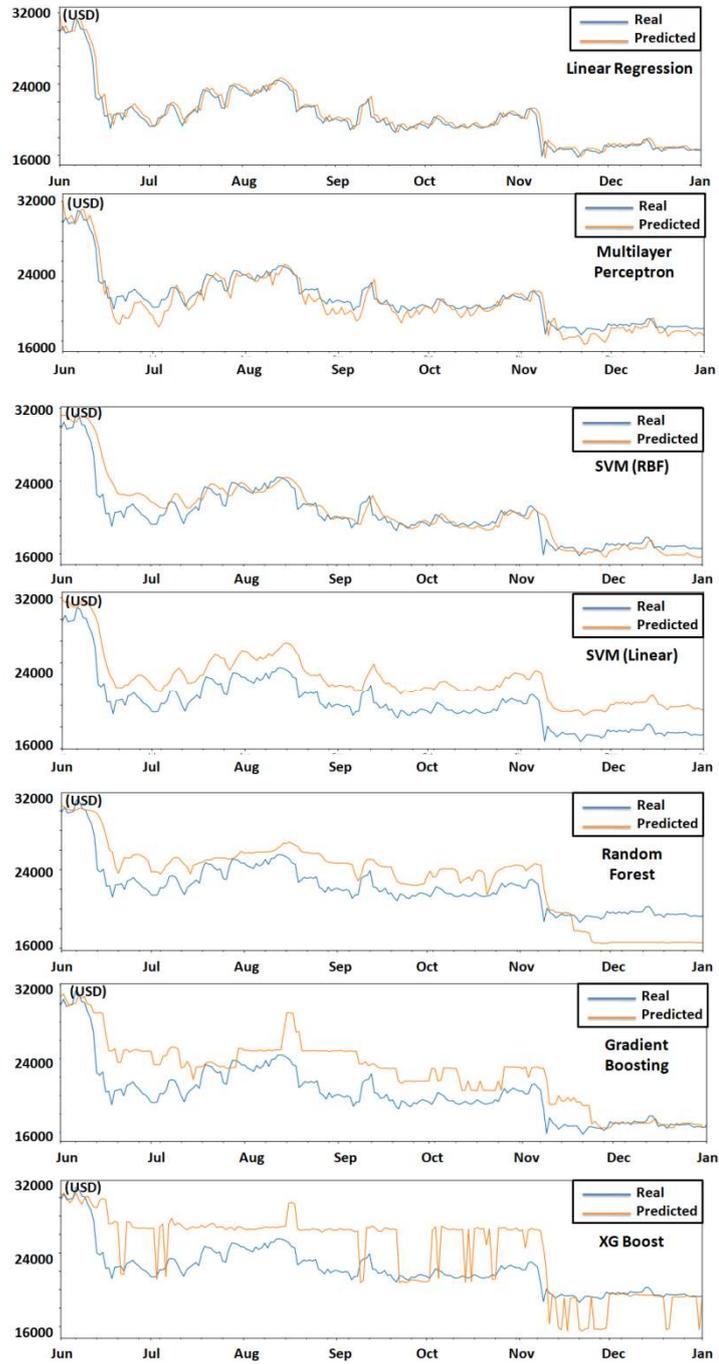

**Figure 4.** Investment simulation predictions considering each of the Machine Learning algorithms.